\documentclass[10pt,twocolumn,letterpaper]{article}

\usepackage[pagenumbers]{cvpr} 

\usepackage{graphicx}
\usepackage{amsmath}
\usepackage{amssymb}
\usepackage{booktabs}
\usepackage{siunitx}
\usepackage{bm}
\usepackage{subcaption}
\usepackage{soul}
\usepackage[dvipsnames]{xcolor}
\usepackage{array}
\usepackage{pifont}
\usepackage[accsupp]{axessibility}

\usepackage[pagebackref,breaklinks,colorlinks]{hyperref}

\usepackage[capitalize]{cleveref}
\crefname{section}{Sec.}{Secs.}
\Crefname{section}{Section}{Sections}
\Crefname{table}{Table}{Tables}
\crefname{table}{Tab.}{Tabs.}

\newcommand{\textpara}[2]{\vspace{0.1in}\noindent\textbf{#1}.~{#2}}
\newcommand{\quespara}[2]{\vspace{0.1in}\noindent\textbf{#1}?~{#2}}

\newcommand{\naq}{\textbf{\texttt{\small NaQ}} }
\newcommand{\naqpp}{\textbf{\texttt{\small NaQ++}} }
\newcommand{\naqb}{\textbf{\texttt{NaQ}} }
\newcommand{\tb}[1]{\textbf{#1}}
\newcommand{\tbg}[1]{\textbf{\color{ForestGreen}{#1}}}
\newcommand{\tbr}[1]{\textbf{\color{red}{#1}}}

\newcolumntype{P}[1]{>{\centering\arraybackslash}p{#1}}
\newcommand{\rcol}[1]{\multicolumn{1}{r}{#1}}
\newcommand{\mctwo}[1]{\multicolumn{2}{c}{#1}}
\newcommand{\ftext}[1]{{\footnotesize #1}}
\newcommand{\cmark}{\ding{51}}%
\newcommand{\xmark}{\ding{55}}%

\definecolor{orange}{RGB}{255,127,0}

\def\confName{CVPR}
\def\confYear{2023}

\begin{document}

\title{NaQ: Leveraging Narrations as Queries to Supervise Episodic Memory}

\author{\textbf{Santhosh Kumar Ramakrishnan$^{1}$~~~~~~Ziad Al-Halah$^2$~~~~~~Kristen Grauman$^{1,3}$} \vspace*{0.05in}\\
$^1$UT Austin~~~~~$^2$University of Utah~~~~~$^3$FAIR, Meta AI
}
\maketitle

\begin{abstract}
Searching long egocentric videos with natural language queries (NLQ) has compelling applications in augmented reality and robotics, where a fluid index into everything that a person (agent) has seen before could augment human memory and surface relevant information on demand. However, the structured nature of the learning problem (free-form text query inputs, localized video temporal window outputs) and its needle-in-a-haystack nature makes it both technically challenging and expensive to supervise. We introduce Narrations-as-Queries (NaQ), a data augmentation strategy that transforms standard video-text narrations into training data for a video query localization model.  Validating our idea on the Ego4D benchmark, we find it has tremendous impact in practice.  NaQ improves multiple top models by substantial margins (even doubling their accuracy), and yields the very best results to date on the Ego4D NLQ challenge, soundly outperforming all challenge winners in the CVPR and ECCV 2022 competitions and topping the current public leaderboard. Beyond achieving the state-of-the-art for NLQ, we also demonstrate unique properties of our approach such as the ability to perform zero-shot and few-shot NLQ, and improved performance on queries about long-tail object categories. Code and models: {\small\url{http://vision.cs.utexas.edu/projects/naq}}.
\end{abstract}

\section{Introduction}

Human memory can fail us in day-to-day things in our visual experience.  We misplace objects in the house (\emph{where is my passport?}), we lose track of what tasks we have or have not done (\emph{did I add the salt already?}),  we forget where we did a given activity (\emph{where did I buy tickets last time?}), we do not notice the state of an object in our environment (\emph{did I leave the garage door open?}).  First-person or ``egocentric" perception on a wearable camera could reduce that cognitive overload and provide us with a \emph{superhuman personal episodic memory}---by seeing what we see, and indexing it in meaningful and easy-to-access ways.  

\begin{figure}[t]
    \centering
    \includegraphics[width=0.48\textwidth,clip,trim={0 8.3cm 18.5cm 0}]{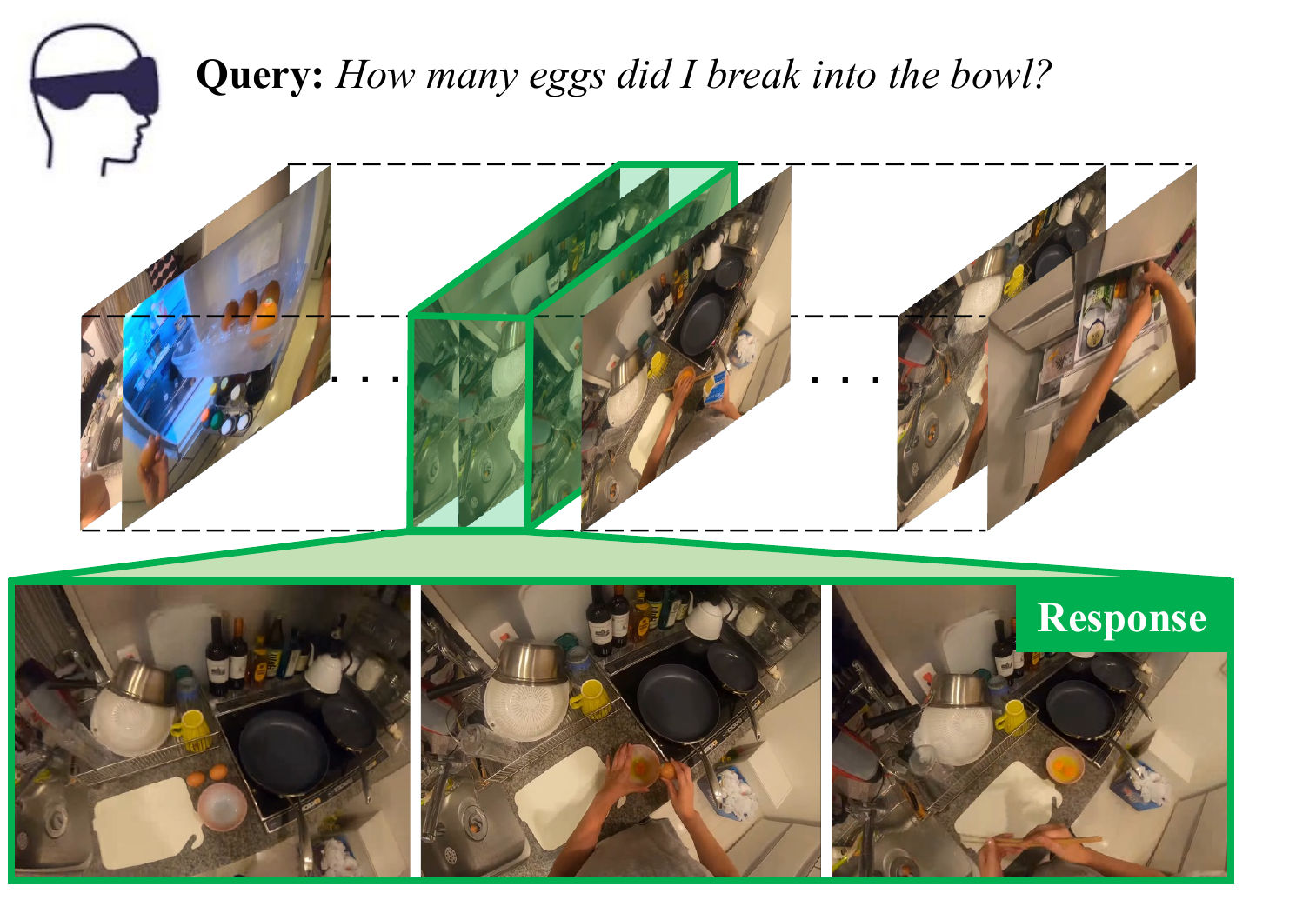}
    \vspace*{-0.3in}
    \caption{ 
    Episodic memory with natural language queries (NLQ) aims to search long egocentric videos to identify the temporal ``response" window revealing the answer to a free-form question about the camera wearer's past visual experience.
    }
    \label{fig:intro}
    \vspace{-0.3cm}
\end{figure}

This is the vision of the Natural Language Query (NLQ) task in Ego4D's Episodic Memory benchmark~\cite{grauman2022ego4d}.  Given a natural language question and a long egocentric video, the NLQ task requires identifying the precise temporal window in the camera wearer's past video that reveals the answer.  See Figure~\ref{fig:intro}.  Such functionality could transform the everyday experience of an augmented reality user with always-on AR glasses.  It could similarly play a role for a mobile household robot, whom a user may wish to query about its 
visual history (\emph{have you seen my keys?}).

The NLQ challenge has attracted substantial attention in the research community over the last year~\cite{zheng2022exploring,lin2022egocentric,liu2022reler} as have related video-language efforts for question answering~\cite{xu2017video,rohrbach2017movie,xu2021vlm,zhang2020span,zhang2020learning,yang2021just}. The technical challenges are striking.  Queries are free-form natural language, response windows are tiny slivers (a few seconds or less) within a long stretch of video, and wearable camera video is notoriously noisy with its quick head motions and limited field of view.  

\begin{figure}[t]
    \centering
    \includegraphics[width=0.45\textwidth,clip,trim={0 7cm 17cm 0}]{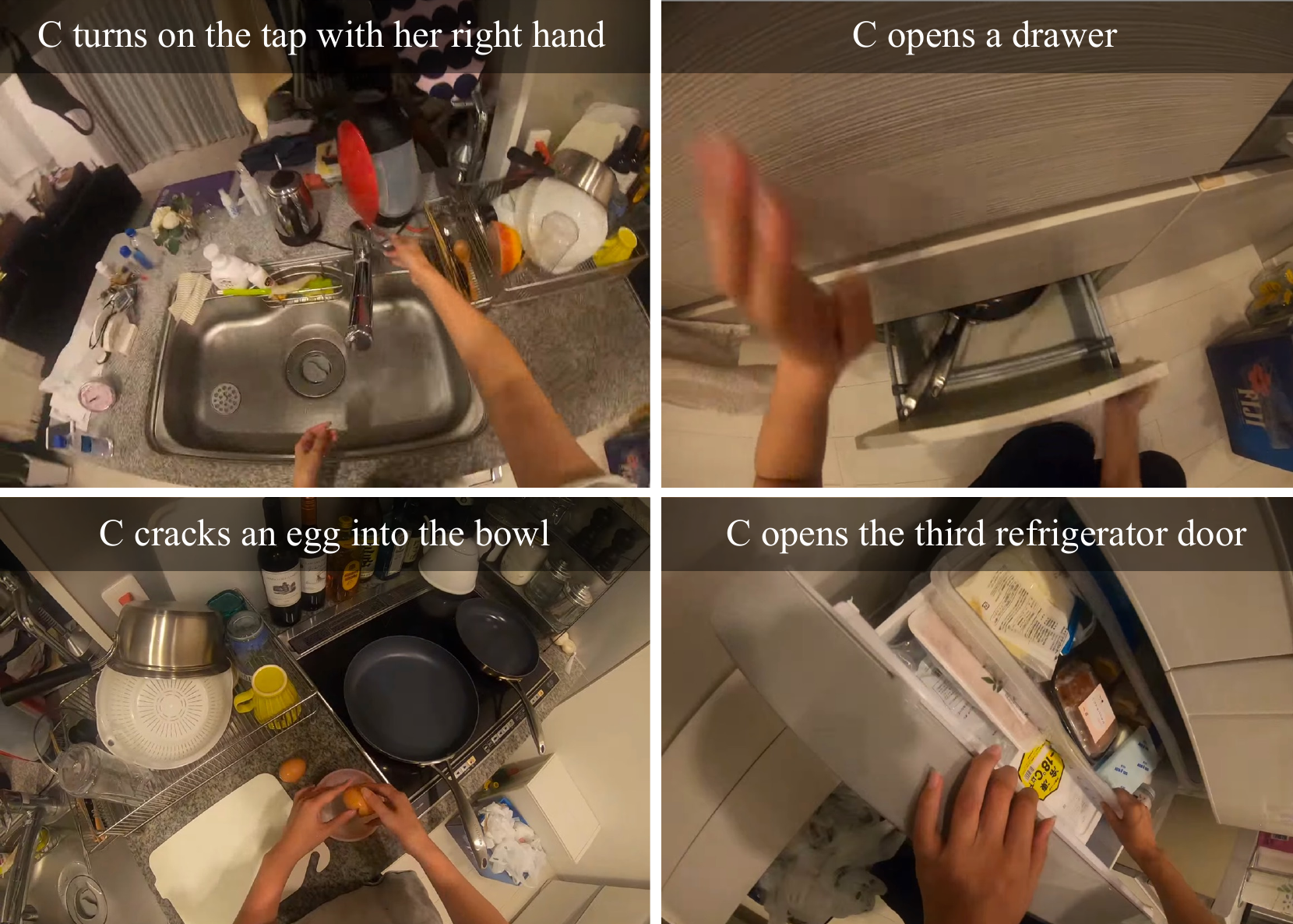}
    \vspace{-0.1in}
    \caption{\textbf{Narration examples.} ``C" refers to the camera-wearer.}
    \label{fig:narrations}
    \vspace{-0.3cm}
\end{figure}

Today's most successful methods embrace the visual-language aspect of the problem.  In particular, inspired by the growing success of visual-linguistic embeddings~\cite{sun2019videobert,li2020hero,radford2021learning,miech2020end,yang2021just}, top competitors on NLQ perform large-scale pretraining on $\langle$video clip, text description$\rangle$ pairs mined from the Ego4D dataset's provided \emph{narrations}~\cite{lin2022egocentric}, which are timestamped play-by-play descriptions of the camera-wearer's activity (see Figure~\ref{fig:narrations}).  The result is a video backbone enhanced by the semantics of grounded language.

While it is important to have strong video and text representations, the downstream \emph{query localization models} that search the video for a response are also crucial to NLQ, yet relatively starved for data.  This is a direct consequence of the difficulty in annotating a query-response pair (which entails posing a creative question and scrolling the long video to mark the temporal response window) versus the relative ease in narrating a video (which entails pausing the video at regular intervals and describing what happened).  For example, whereas Ego4D has 3,670 hours of data annotated with narrations---more than 3.85M sentences in total---it offers only 227 hours of NLQ query examples, for 19k total text queries.  Accordingly, existing methods risk failing to learn about things that are poorly represented in training, such as queries about objects in the long-tail or complex queries involving interactions between multiple visual entities.

To address this issue, we introduce Narrations-as-Queries (NaQ), a simple but exceptionally effective data augmentation strategy for NLQ. NaQ is a novel strategy that uses timestamped narrations to expand the supervision available for training query-localization modules within an episodic memory architecture. Our hypothesis is that narrations provide descriptive information that is localizable in long videos, and thus can benefit an episodic memory model when used as training queries. Specifically, we derive $\langle$video, language query, temporal window response$\rangle$ annotations from timestamped narrations, and augment the conventional query-response data with these pseudo-queries.  Importantly, this allows us to influence the localization module---the workhorse responsible for finding a needle in a haystack---with multimodal data, as opposed to just the video and text encoders. 

Empirically, our idea has tremendous impact. Demonstrating NaQ on the Ego4D Episodic Memory benchmark, we find our simple augmentation strategy successfully complements multiple existing state-of-the-art episodic memory methods, achieving sizeable improvements  (e.g., 32\% to 125\% relative jumps in accuracy) across query types, metrics, and methods. Notably, our gains hold even compared to existing methods such as EgoVLP~\cite{lin2022egocentric} that use the same (or even more) narration annotations as our model, meaning that NaQ's success can be attributed to good modeling, not more data. Moreover, NaQ even benefits video-language grounding on exocentric videos, i.e., it is beneficial to augment its exocentric training with narrated egocentric videos. To our knowledge, NaQ yields the very best results to date on the NLQ challenge, strongly outperforming all the challenge winners from Ego4D CVPR'22 and ECCV'22, and topping the current public leaderboard. Beyond achieving state-of-the-art results, we perform a thorough analysis of the strengths and weaknesses of NaQ, and demonstrate useful properties such as benefits on long-tail object queries as well as zero-shot and few-shot NLQ. 

\section{Related work}

\textpara{Egocentric video understanding}{
Prior work has developed video datasets and methods for egocentric perception~\cite{grauman2022ego4d,Damen2022RESCALING,fathi2011gtea,kazakos2019epic,furnari2020rolling}. Egocentric video offers a camera wearer's perspective of their activities over a long time horizon and raises challenging research problems such as human-object interactions~\cite{cai2018understanding,damen2014you}, activity recognition~\cite{kazakos2019epic,zhou2015temporal}, anticipation~\cite{abu2018will,girdhar2021anticipative}, episodic memory~\cite{grauman2022ego4d}, and video summarization~\cite{del2016summarization,yongjae-ijcv2015}. In this work, we tackle the episodic memory task.  We leverage the Ego4D dataset~\cite{grauman2022ego4d}, which consists of 3,670 hours of video of daily-life activity captured by 931 camera wearers around the world.
}\vspace*{-0.05in}

\textpara{Vision-language pretraining (VLP)}{VLP methods  rely on large-scale video-text datasets~\cite{miech2019howto100m,bain2021frozen} to learn transferable representations for video-language tasks such as retrieval~\cite{hendricks2018localizing,escorcia2019temporal}, question-answering~\cite{rohrbach2017movie,xu2021vlm} and video captioning~\cite{krishna2017dense,zhou2018end}. VideoBert learns joint video-text embeddings by discretizing video frames and performing BERT-like pre-training~\cite{sun2019videobert}. HERO improves over this with a hierarchical encoding of multi-modal inputs~\cite{li2020hero}. MIL-NCE learns to match clips with temporally close captions to address video-text misalignment in HowTo100M~\cite{miech2020end,miech2019howto100m}. While these methods primarily focus on third-person videos, EgoVLP~\cite{lin2022egocentric} adapts the InfoNCE objective to egocentric settings and uses video-narration annotations from Ego4D~\cite{grauman2022ego4d} to learn video-text backbones for ego-video understanding. Just-Ask~\cite{yang2021just} proposes a strategy to generate video question-answering data consisting of (short clips, questions, text answers) from narrated YouTube videos. 

While we take inspiration from such methods, our idea is very different. Unlike prior work that learns representations or video-QA systems from short video clips and (possibly weakly) aligned text, we learn to \emph{temporally localize} short text queries in long untrimmed videos egocentric videos. Whereas Just-Ask's data generation procedure~\cite{yang2021just} outputs questions with \emph{text} responses for short video clips, ours outputs temporal windows in long videos. Rather than pretraining a video/text backbone~\cite{lin2022egocentric,sun2019videobert,li2020hero,miech2020end}, our model injects multimodal supervision to train a query-localization module.  Overall, our idea is complementary to prior video-text pretraining efforts, as we will demonstrate in the results. 
}\vspace*{-0.05in}

\textpara{Video-language grounding}{
Prior work performs video-language grounding (VLG) in exocentric videos~\cite{zhang2020span,rohrbach2014coherent,gao2017tall,krishna2017dense,zhang2020learning}. The Ego4D episodic memory benchmark first introduced NLQ, a new VLG task requiring temporal query localization in long egocentric videos~\cite{grauman2022ego4d}. Existing VLG methods like 2D-TAN~\cite{zhang2020learning} and VSLNet~\cite{zhang2020span} have been adapted to perform NLQ, while the recent ReLER~\cite{liu2022reler} model achieves state-of-the-art NLQ using a multi-scale and cross-modal transformer with video-level data augmentation. Our goal is to improve such methods via large-scale data augmentation with narration-based queries. In addition, our proposed strategy performs \emph{query-level augmentation} and is complementary to the video-level data augmentation from~\cite{liu2022reler}. Recent work uses point-wise (aka ``glance") annotations to reduce annotation costs for VLG training~\cite{xu2022point,cui2022video}. However, these are limited to exocentric videos and assume \emph{task-specific} point annotations, whereas the Ego4D narrations are not specific to the NLQ task.  As we will demonstrate in experiments, our approach stacks well when combined with prior NLQ methods~\cite{liu2022reler,zhang2020span,lin2022egocentric}, and can even benefit exocentric VLG via an ego-exo transfer of the egocentric narrations.
}

\begin{figure*}[t]
    \centering
    \includegraphics[width=\textwidth,clip]{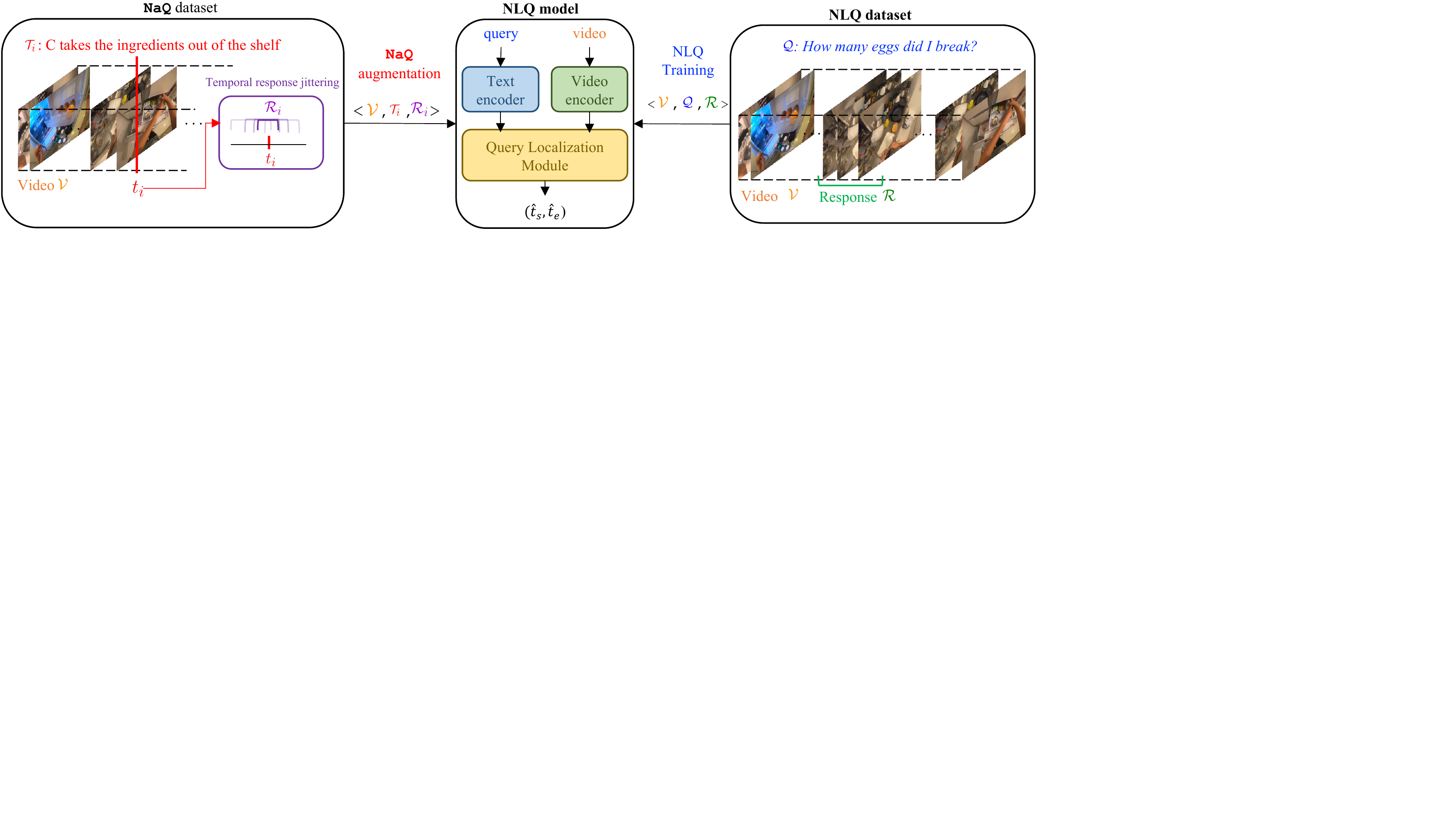}
    \vspace*{-0.15in}
    \caption{\textbf{Narrations-as-Queries:} We propose a simple-yet-effective data-augmentation strategy for natural language queries (NLQ). The status-quo NLQ methods train in a supervised fashion on annotated ($\mathcal{V}$: video, $\mathcal{Q}$: query, $\mathcal{R}$: response) tuples, where the response is a $(t_s, t_e)$ temporal window (see right). This is severely limiting, since such task-specific data is expensive to obtain and is available only on a small scale. We propose a narrations-as-queries pipeline to tackle this issue (see left). Our key idea is to leverage densely annotated video narrations, where each narration $\mathcal{T}_i$ for video $\mathcal{V}_j$ is a textual description of the camera-wearer's activity at time $t_i$. We propose ``temporal response jittering", a technique to convert timestamped narrations into natural language queries with temporal response windows $\langle \mathcal{V}_j, \mathcal{T}_i, \mathcal{R}_i \rangle$ and obtain the \naq dataset, which contains $80\times$ more samples when compared to the NLQ dataset. We then train various NLQ models jointly on the NLQ and \naq datasets to obtain significant gains across query types, architectures, and metrics.
    \vspace*{-0.15in}
    }
    \label{fig:naq_approach}
\end{figure*}

\section{Approach}
Our key insight is to leverage narrations as an additional data source to improve a model's ability to localize answers in a long video when prompted with a natural language query. To do this, we propose a strategy to convert narrations and their timestamps into NLQ annotations. Our strategy is automatic and simple which allows us to scale the training data for episodic memory search by two orders of magnitude.

We first define the episodic memory task (\cref{sec:em_task}), then our Narrations-as-Queries approach to convert narrations into NLQ annotations (\cref{sec:naq_dataset}), and finally our training strategy (\cref{sec:naq_training}).

\subsection{Episodic memory with natural language query}
\label{sec:em_task}
The goal of episodic memory is to perform query-driven reasoning about long-form egocentric videos. First introduced in Ego4D~\cite{grauman2022ego4d}, it is well-motivated by applications discussed above, such as augmented reality assistants that enable superhuman memory. The NLQ task has attracted significant attention in the research community, with 10+ teams around the world competing on the benchmark over the last year~\cite{zheng2022exploring,lin2022egocentric,liu2022reler}, organized challenges at CVPR'22 and ECCV'22, and an active public leaderboard.\footnote{Ego4D NLQ challenge: {\scriptsize \url{https://eval.ai/web/challenges/challenge-page/1629/overview}}}

More formally, given an egocentric video $\mathcal{V}$ capturing a camera wearer's past experiences and a natural language query $\mathcal{Q}$ in the form of a question, the task requires temporally localizing where the answer can be seen in the video, i.e., a response window $\mathcal{R} = [t_s, t_e]$. For example, the query could be $\mathcal{Q}=$``\textit{What vegetables did I put in the soup the last time I made it?}", and the model needs to search a given video $\mathcal{V}$ to identify the time window $[t_s, t_e]$ that contains the answer, i.e., the type of vegetables in the soup. A data sample for this task is of the form $\langle$video, query, response$\rangle$. The video can be several minutes long, and the response to the query can appear in a time window that is shorter than a second, making this a very challenging task.

\subsection{Narrations-as-Queries}
\label{sec:naq_dataset}
Prior NLQ methods are limited in performance due to the lack of large-scale NLQ annotations of the form $\langle$video, query, response$\rangle$.  We address this limitation by proposing a method to automatically transform narrations associated with egocentric videos to a compatible form for NLQ. Narrations are free-form sentences describing the current activity performed by the camera-wearer (see~\cref{fig:narrations}). They are time-stamped and temporally dense (e.g., there are 13.2 sentences per minute of video on average in Ego4D~\cite{grauman2022ego4d}). 

These annotations are substantially cheaper to obtain than NLQ annotations. For narrations, the annotators needs to simply describe the activity that is seen in the video; whereas for NLQ, first a meaningful, unambiguous question needs to be formulated and then the annotator needs to manually search the video back and forth to identify the time window that shows the answer. Hence, narrations can be annotated at a much larger scale compared to NLQ (e.g., Ego4D has 3.85M narrations vs.~19k NLQ samples). Moreover, narrations have several applications beyond NLQ~\cite{lin2022egocentric,chen2022internvideo,ashutosh2023hiervl,nair2022r3m}, and are likely to be invested in on a large-scale. 

Our idea is to leverage this massive data source to aid learning for the NLQ task. We achieve this by first generating a temporal window associated with each narration that approximately captures when the activity described by the narration started and ended. Then, we use these samples (narrations coupled with temporal windows) as additional supervision to train an NLQ localization model to identify where these narrations happen in the video (see~\cref{fig:naq_approach}). Next, we formally describe our approach in detail.
\vspace{-0.05in}

\textpara{1. Generating temporal windows for narrations}{
Each video narration consists of a textual sentence $\mathcal{T}$, and a single timestamp $t$ marking the correspondence to the underlying video (see ~\cref{fig:naq_approach}, left). However, this is incompatible with NLQ task architectures which require queries and temporal response windows as supervision. To address this, we propose \emph{temporal response jittering}, a technique to convert narration timestamps to temporal windows conditioned on the video. \vspace{0.05in}

\textbf{Temporal response jittering:} Our goal is to convert a narration timestamp $t_i$ from video $V_j$ into a response window $\mathcal{R}_i=(t_s, t_e)$. First, we use ``contextual variable-length clip pairing strategy" introduced in EgoVLP~\cite{lin2022egocentric} to obtain a \emph{video-conditioned} seed temporal window centered around $t_i$:\vspace{-0.10in}
\begin{equation}
    \bar{\mathcal{R}_i} = [t_i - \beta_j / 2\alpha, t_i + \beta_j / 2\alpha]
    \label{eqn:seed}
\end{equation}
where $\beta_j$ captures the average temporal length between consecutive narrations in video $V_j$, and $\alpha$ is the average of all $\beta_j$ across all videos (please see~\cite{lin2022egocentric} for details). While this offers a good starting point, it fails to address the inherent noise in $\bar{\mathcal{R}_i}$ arising from the lack of explicit human annotation. The responses generated are also typically short (less than a second) and do not match the distribution over NLQ response windows that are 10 seconds long on average. To account for these factors, we transform $\bar{\mathcal{R}_i} = (\bar{t}_s, \bar{t}_e)$ further using a randomized window expansion and translation:
\begin{equation}
    \mathcal{R}_i = [(\bar{t}_c - \delta_t) - s\Delta , (\bar{t}_c - \delta_t) + s\Delta],
    \label{eqn:trj}
\end{equation}
where $\Delta = (\bar{t}_e - \bar{t}_s) / 2$ is the half-width of $\mathcal{\bar{R}}_i$, $\bar{t}_c = (\bar{t}_s + \bar{t}_e) / 2$ is the center of $\mathcal{\bar{R}}_i$, $s \sim U[1, S]$ is an expansion factor, and $\delta_t \sim U[-T, T]$ is a translation factor. Intuitively, the translation factor $\delta_t$ randomly shifts $\bar{\mathcal{R}}$ to model uncertainty in its estimate, and the scaling factor $s$ randomly expands $\bar{\mathcal{R}}$ to match the distribution of NLQ response windows. $S$ is a hyperparameter selected through validation, and $T$ is set as $(s - 1)\Delta$ after sampling $s$ to ensure that the seed temporal window $\bar{\mathcal{R}_i}$ is contained within $\mathcal{R}_i$.

Following this strategy, we can extract narrations and their inferred temporal windows for all video clips with available narrations (denoted by $\mathcal{V}$) to obtain a dataset 
\begin{equation}
  \mathcal{D} = \big\{ (\mathcal{N}_1^v, \cdots, \mathcal{N}_n^v)~~|~~\forall v \in \mathcal{V}\big\},
\end{equation}
where $\mathcal{N}_i^v = \big(\mathcal{T}_i, \mathcal{R}_i \big)$ is the transformed sample that consists of a narration and its corresponding response window. We apply this method to the video clips from the train split of the Ego4D Episodic Memory benchmark to create a dataset $\mathcal{D}$ that contains $850\textrm{k}$ samples of transformed narrations from 4,851 video clips. 
}

\textpara{2. Generating episodic memory queries}{
Given the previous dataset of narrations with associated temporal windows $\mathcal{D}$, we now convert these to a dataset of NLQ queries. Specifically, given a video $\mathcal{V}_j$, we sample a narration $\mathcal{N}_i$ from $\mathcal{V}_j$ and obtain the task input $X = (\mathcal{V}_j, \mathcal{T}_i)$, where $\mathcal{T}_i$ is the narration text, and the label $Y = \mathcal{R}_i$ which represents the start and end times for a narration as defined in~\cref{eqn:trj}. In other words, the narration $\mathcal{T}_i$ becomes the query that effectively asks the model to locate in $\mathcal{V}_j$ where the activity described by $\mathcal{T}_i$ can be found, i.e., the response window $(t_{i}^{start}, t_{i}^{end})$. We found that simply using narration text as the query to work well. This can be attributed to pretrained BERT query encoders used in NLQ models~\cite{zhang2020span,lin2022egocentric,liu2022reler}, which can effectively adapt to the difference between declarative sentences and questions. However, it would be interesting to study techniques to transform narrations to questions in future work~\cite{yang2021just}. This dataset of $(X, Y)$ pairs is our Narrations-as-Queries (\naq) dataset. Next, we incorporate this dataset into the NLQ training pipeline as a form of data augmentation.
}

\subsection{Narrations-as-Queries training for NLQ}
\label{sec:naq_training}
Our \naq is model-agnostic: it stands to benefit any NLQ model out of the box without any model-specific modifications. We demonstrate the universal advantage of \naq by benchmarking several baselines with \naq in experiments.

Specifically, for a given NLQ model $\mathcal{M}$, we train it with \naq in two stages. Let us denote the \naq dataset as $\mathcal{D}_{\textrm{NaQ}}$ and the NLQ train dataset as $\mathcal{D}_{\textrm{NLQ}}$. First, we jointly train $\mathcal{M}$ with both $\mathcal{D}_{\textrm{NaQ}}$ and $\mathcal{D}_{\textrm{NLQ}}$, effectively treating \naq as a query augmentation strategy.  Since \naq expands the training dataset significantly (by 2 orders of magnitude in size), we rely on large batch training with 2048 batch size and an appropriately large initial learning rate of 0.001 on 4-8 A40 GPUs. We train in this large-batch setting for 200 epochs, with early stopping when the validation performance saturates. We then finetune the model on $\mathcal{D}_{\textrm{NLQ}}$ with the default small-batch training used for $\mathcal{M}$, and perform a grid search to determine the learning rate based on $\mathcal{M}$ performance on the validation split.  

\section{Experiments}
\label{sec:expt_setup}

We evaluate our approach on the NLQ task from the episodic memory benchmark from Ego4D~\cite{grauman2022ego4d}. This benchmark has gained significant interest and has been the subject of two Ego4D challenges held at CVPR 2022 and ECCV 2022. The NLQ task contains $11.3$k / $3.9$k / $4$k queries annotated over $136 / 45 / 46$ hours of train / val / test videos. Each video clip is $8.2$ minutes on average, and the ground-truth query response is $10.5$ seconds on average in the train dataset. That means the response window occupies only 2\% of the input video on average. We primarily perform experiments on Ego4D since it is consistent with our episodic memory motivation and uniquely supports our setting with a combination of $\langle$\emph{egocentric videos, NLQ annotations, large-scale narrations}$\rangle$. We additionally experiment on the TACoS dataset of exocentric kitchen videos to test the generalization of our approach~\cite{rohrbach2014coherent}. It contains $10$k / $4.5$k queries annotated over $75 / 25$ train / val videos, and offers long videos with short response windows. 

\textpara{Evaluation metrics}{We measure performance on NLQ using metrics from the video-language grounding literature and adapted for NLQ in~\cite{grauman2022ego4d}. We report the recall@k, IoU=m metric, where $\textrm{k}$ = $\{1, 5\}$ and $\textrm{m}$ = $\{0.3, 0.5\}$. This measures the percentage of times where at least one of the top-k predicted candidates have at least an intersection-over-union (IoU) of m.
}

\textpara{Baselines}{
We evaluate the impact of \naq by combining it with 3 existing methods in the literature.

\textbf{(1) VSLNet} treats natural-language grounding as a text-based question answering problem~\cite{zhang2020span}. It represents the input video as a text passage and uses a span-based QA framework~\cite{seo2016bidirectional} to localize responses to text queries. This was adapted to perform the NLQ task in~\cite{grauman2022ego4d} by using SlowFast features pretrained on Kinetics 400~\cite{feichtenhofer2019slowfast}.

\textbf{(2) EgoVLP} proposes to pretrain video and text backbones on the EgoNCE pretraining task~\cite{lin2022egocentric}. By leveraging large-scale video + text narrations from Ego4D, they successfully transfer features to a variety of tasks including NLQ. It was the runner-up entry for the Ego4D NLQ challenge at CVPR 2022. This method replaces the SlowFast features from the VSLNet baseline with the EgoVLP pretrained backbones. This baseline is complementary to our own approach where we use narrations to augment the localization training for the NLQ task.  

\textbf{(3) ReLER} adapts VSLNet to use a multi-scale cross-modal transformer architecture~\cite{liu2022reler}. It also proposes to augment the training data using video-level augmentation strategies like randomly sampling a subset of the video to try and mitigate overfitting. This was the winning entry of the Ego4D NLQ challenge at CVPR 2022. We augment ReLER with EgoVLP pretrained backbones to obtain a stronger `ReLER$^*$' baseline. Unlike ReLER, which augments the data at the video level, we propose to augment the data at the query level. We will demonstrate that \naq is complementary and boosts the performance of ReLER. 

Note that \naq leverages the same narrations as EgoVLP and ReLER$^*$, and requires no greater supervision or data. 
}

\textpara{Implementation details}{
For each baseline, we adapt the authors' code to train with \naq data augmentation. For consistency, we report the results of each method as reproduced using the provided code, in addition to reporting the official paper numbers. We train each method with \naq augmentation for 200 epochs and stop training early when the validation performance saturates. We found that it was helpful to finetune for up to 30 epochs on only the NLQ dataset. Please see~\cref{suppsec:implementation} in supp. for details.
}

\begin{table}[t]
\centering
\resizebox{0.48\textwidth}{!}{
\begin{tabular}{@{}clccccc@{}}
\toprule
       &                                            &           & \multicolumn{2}{c}{IoU=0.3}     & \multicolumn{2}{c}{IoU=0.5} \\ 
       &  \multicolumn{1}{c}{Method}      & {\small Narrations} &       R@1      &     R@5        &     R@1      &    R@5       \\ \midrule
  1.   &  VSLNet~\cite{zhang2020span}               &  \xmark   &      5.45      &     10.74      &    3.12      &    6.63      \\
  2.   &  VSLNet$^\dagger$                          &  \xmark   &      5.21      &     11.19      &    2.78      &    6.72      \\
  3.   &  VSLNet + \naq                             &  \cmark   &   \tb{10.26}   & \tb{19.01}     &\tb{5.81}     &\tb{12.67}    \\ 
       &  \multicolumn{1}{r}{\small absolute gain}  &           & \tbg{+5.05}    & \tbg{+7.82}    &\tbg{+3.03}   &\tbg{+5.95}    \\
    \midrule
  4.   &  EgoVLP~\cite{lin2022egocentric}           &  \cmark   &      10.84     &     18.84      &    6.81      &    13.45     \\ 
  5.   &  EgoVLP$^\dagger$                          &  \cmark   &      10.40     &     19.33      &    6.18      &    13.03     \\ 
  6.   &  EgoVLP + \naq                             &  \cmark   &   \tb{15.90}   & \tb{26.38}     &\tb{9.46}     &\tb{17.80}    \\ 
       &  \multicolumn{1}{r}{\small absolute gain}  &           & \tbg{+5.50}    &\tbg{+7.05}     &\tbg{+3.28}   &\tbg{+4.77}    \\ 
    \midrule
  7.   &  ReLER~\cite{liu2022reler}                 &  \xmark   &      10.79     &     13.19      &    6.74      &     8.85     \\ 
  8.   &  ReLER$^\dagger$                           &  \xmark   &       9.91     &     12.29      &    6.17      &     8.03     \\ 
  9.   &  ReLER$^*$                                 &  \cmark   &      14.66     &     17.84      &    8.67      &    11.54     \\ 
  10.  &  ReLER$^*$ + \naq                          &  \cmark   &   \tb{19.31}   & \tb{23.62}     &\tb{11.59}    &\tb{15.75}    \\
       &  \multicolumn{1}{r}{\small absolute gain}  &           & \tbg{+4.65}    &\tbg{+5.78}     &\tbg{+2.92}   &\tbg{+4.21}    \\
\bottomrule
\end{tabular}
}
\vspace{-0.25cm}
\caption{\textbf{Results on Ego4D NLQ dataset.} $^*$replace SlowFast with EgoVLP features. $^\dagger$Results reproduced using authors' code.}
\label{tab:nlq-results}
\vspace{-0.25cm}
\end{table}

\subsection{Experimental results on Ego4D NLQ}

We report results on the NLQ validation set in~\cref{tab:nlq-results}. The poor performance of the VSLNet baseline on NLQ highlights the difficulty of the task. It requires localizing responses typically shorter than 10 seconds in 8+ minute long egocentric videos. The limited size of the training dataset further exacerbates this problem, since there are only 11.3k training queries. However, when augmented with \naq, the performance across all metrics nearly doubles, indicating the effectiveness of \naq in addressing these challenges. This is a dramatic gain, though it comes at the cost of larger narrations data that is not available to VSLNet.

When VSLNet is augmented with \naq, it is already competitive with EgoVLP, which pretrains video and text backbones with Ego4D videos + narrations and uses the same VSLNet query-localization architecture (rows 3 vs.~5). When \naq is combined with EgoVLP, it further improves the performance by 3.2 - 7.1 points across metrics (rows 5 vs.~6). This confirms that \naq augmentation for query localization training complements the EgoVLP pretraining of video-text backbones.  Importantly, our gain here comes at no additional cost in data or annotations.

ReLER~\cite{liu2022reler} uses SlowFast + CLIP video features. For a fair comparison, we replace the SlowFast features with EgoVLP features to obtain ReLER$^{*}$. This improves by a large margin as expected, and gives us a stronger baseline to compare with (rows 8 vs.~9). Recall that ReLER$^*$ uses video-level data augmentation using variable-length sliding windows and video splicing~\cite{liu2022reler}. When ReLER$^*$ is augmented with \naq, the performance increases by a significant margin. This confirms the complementary nature of the query-level augmentation we propose in \naq with video-level augmentation in ReLER.

Overall, we find that \naq augmentation greatly improves the performance of all methods across all metrics. The absolute gains across metrics are remarkably consistent regardless of the underlying method. When averaged across the methods, \naq improves the absolute recall@1 performance by $5.06$ at IoU=0.3 and $3.08$ at IoU=0.5, and the absolute recall@5 performance by $6.88$ at IoU=0.3 and $4.98$ at IoU=0.5. This confirms the generality and effectiveness of \naq at expanding the limited NLQ annotations by bootstrapping it with narrations, a relatively cheaper and more abundant data source. More importantly, the insight in \naq is not simply that large-scale data benefits performance. Rather, we emphasize \emph{how} to use this data: we leverage \emph{narrations as queries} for query-localization network training. This is evidenced by our experiments demonstrating major gains on EgoVLP and ReLER$^*$, methods which also benefit from large-scale pretraining on video-narrations data.  \vspace{0.1in}

\begin{table}[t]
\centering
\resizebox{0.49\textwidth}{!}{
\begin{tabular}{@{}lP{1cm}P{1cm}P{1cm}P{1cm}P{1cm}@{}}
\toprule
\multicolumn{1}{c}{Method}          & R@1 IoU=0.3    &     R@1 IoU=0.5  & Mean R@1$^\dagger$ & R@5 IoU=0.3  & R@5 IoU=0.5       \\ \midrule
  \naqpp~(ours)$^\ddagger$          &   \tbg{21.70}  &  \tbg{13.64}     &  \tbg{17.67}       &      25.12   &      16.33         \\ \midrule
  \naq~(ours)                       &   \tb{18.46}   &  \tb{10.74}      &  \tb{14.59}        &      21.50   &      13.74         \\
  InternVideo~\cite{chen2022internvideo} &  16.46    &      10.06       &      13.26         &      22.95   &      16.11         \\
  Badgers@UW-Mad.~\cite{mo2022simple}&      15.71    &       9.57       &      12.64         &  \tb{28.45}  &  \tb{18.03}        \\
  CONE~\cite{hou2022efficient}      &       15.26    &       9.24       &      12.25         &      26.42   &      16.51         \\
  ReLER~\cite{liu2022reler}         &       12.89    &       8.14       &      10.51         &      15.41   &      9.94          \\
  EgoVLP~\cite{lin2022egocentric}   &       10.46    &       6.24       &      8.35          &      16.76   &      11.29         \\
  VSLNet~\cite{zhang2020span}       &       5.42     &       2.75       &      4.08          &      8.79    &      5.07          \\
\bottomrule
\end{tabular}
}
\vspace{-0.25cm}
\caption{\textbf{Results on Ego4D NLQ challenge.} $^\dagger$Primary metric for the challenge. $^\ddagger$Our leaderboard entry post CVPR '23 acceptance.}
\label{tab:nlq-challenge-results}
\vspace{-0.5cm}
\end{table}

\vspace{-0.3cm}
\textpara{Ego4D NLQ challenge}{
We submitted our best performing method (ReLER$^{*}$ + \naq) to the Ego4D NLQ challenge leaderboard, where the NLQ evaluation is performed on an EvalAI server on a held-out set of test annotations~\cite{grauman2022ego4d}. Note that while the videos are available to participants, the annotations (including narrations) are not accessible. The results are shown in~\cref{tab:nlq-challenge-results}. VSLNet is the baseline provided by the organizers. ReLER and EgoVLP were the winning and runner-up entries from the CVPR 2022 edition of the challenge.  InternVideo~\cite{chen2022internvideo}, Badgers@UW-Madison~\cite{mo2022simple}, and CONE~\cite{hou2022efficient} are the top three entries from the ECCV 2022 edition of the challenge. At the time of submission, \naq was the leading entry among all methods on the leaderboard. Post-acceptance, we combined \naq with the ECCV and CVPR challenge winners (i.e., ReLER architecture with InternVideo features) to obtain \naqpp. Our results set the state-of-the-art for NLQ, outperforming prior work by a large margin. \naqpp is also the official baseline for the CVPR 2023 Ego4D NLQ challenge.
}

\textpara{TRJ ablation}{We study the impact of using temporal response jittering (TRJ) (\cref{sec:naq_dataset}) in an ablation study. We observe that using TRJ improves the performance by up to 0.8 points in recall @ 1 metrics and 1.6 in recall @ 5 metrics consistently across all methods. Please see~\cref{suppsec:ablations} for the complete results.}

\subsection{Experimental results on TaCOS NLQ}

Existing third-person (aka exo) video datasets for language grounding lack large-scale narrations, which prevents a direct analogue of our experiments in exo videos. Therefore, we perform an ego-exo variant using the TACoS dataset of exo kitchen videos~\cite{rohrbach2014coherent}, where we jointly train on the \naq dataset from Ego4D's $\langle$ego videos, narrations$\rangle$ and $\langle$exo videos, language queries$\rangle$ from TACoS. See~\cref{tab:tacos}. \naq benefits both VSLNet and EgoVLP. EgoVLP underperforms VSLNet since its video features were pretrained on ego videos, while VSLNet uses SlowFast features pretrained on Kinetics 400. Though the performance gains from \naq are lower than on Ego4D due to the ego/exo domain mismatch, these results reinforce our method's generality.

\begin{table}[t]
\centering
\hspace*{-0.05in}\resizebox{0.50\textwidth}{!}{%
\begin{tabular}{@{}ccccccccc@{}}
\toprule
               & \multicolumn{4}{c}{VSLNet}                                  & \multicolumn{4}{c}{EgoVLP}                                  \\ \cmidrule(lr){2-5}\cmidrule(lr){6-9}
               & \multicolumn{2}{c}{IoU = 0.3} & \multicolumn{2}{c}{IoU=0.5} & \multicolumn{2}{c}{IoU = 0.3} & \multicolumn{2}{c}{IoU=0.5} \\
    \naq       & R@1           & R@5          &  R@1          & R@5          & R@1           & R@5          &  R@1          & R@5          \\ \cmidrule(r){1-1}\cmidrule(lr){2-5}\cmidrule(lr){6-9}
    \xmark     &      20.10    &     29.10    &    15.42      &     22.85    &      16.52    &     25.06    &    12.73      &     19.33    \\
    \cmark     &  \tb{23.39}   & \tb{32.69}   &\tb{19.13}     & \tb{26.43}   &  \tb{18.24}   & \tb{27.25}   &\tb{13.78}     & \tb{20.03}   \\
    \multicolumn{1}{r}{\small abs. gain} &  \tbg{+3.29}   &  \tbg{+3.59}  &   \tbg{+3.71}  &  \tbg{+3.58}  &  \tbg{+1.72}   &  \tbg{+2.19}  &  \tbg{+1.05}   &  \tbg{+0.70}  \\
\bottomrule
\end{tabular}
}
\vspace*{-0.1in}
\caption{Results on TACoS dataset of third-person cooking videos.}
\label{tab:tacos}
\vspace*{-0.1in}
\end{table}

\begin{figure*}[t]
    \centering
    \includegraphics[width=\textwidth,clip]{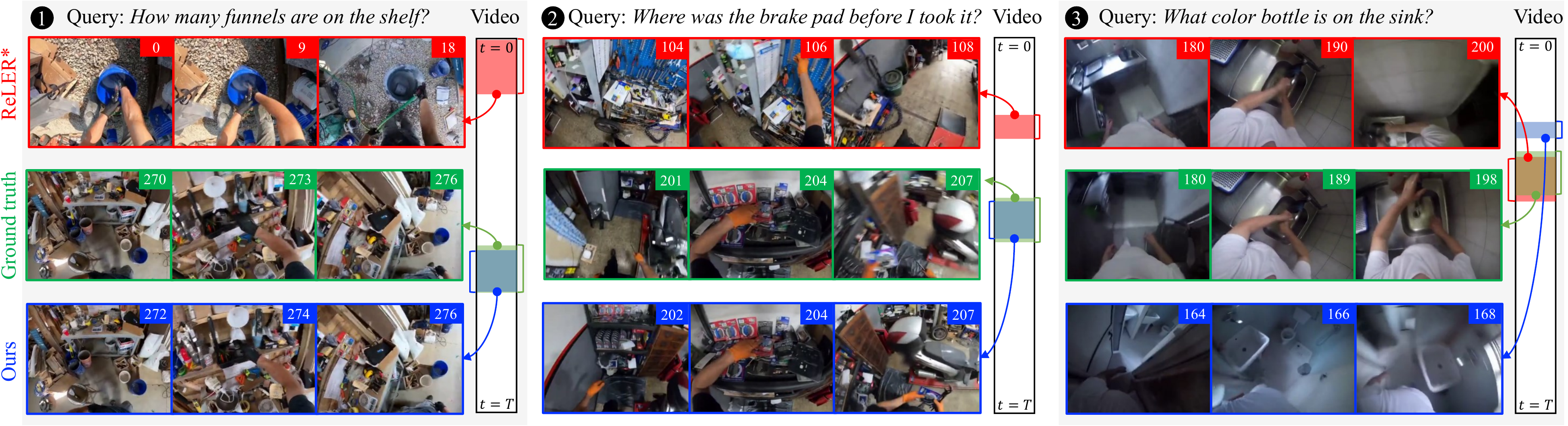}
    \vspace*{-0.25in}
    \caption{\textbf{Qualitative analysis.} We show three examples of NLQ task predictions (one per column). In each column, the natural language query is displayed at the top, the ground truth responses are in the central row, and the model predictions are on the first and last rows. The temporal extents of the video and predicted time windows are shown right next to the images on each column. We compare ReLER$^*$~\cite{liu2022reler} baseline (on the first row) against our \naq method which augments the NLQ training for ReLER$^*$. \textbf{Example~1:} Our method successfully identifies the response window showing how many funnels are on the shelf, while the baseline fails. The object `funnel' is a low-shot object with fewer than 10 training queries. This supports our experimental observation that \naq has a strong advantage on low-shot objects and counting-based queries (see~\cref{tab:result_object_types,tab:result_query_types}). \textbf{Example 2:} \naq successfully recognizes the object `brake pad' and is able to localize where it was taken. ReLER* incorrectly identifies a spanner as the response. \textbf{Example 3:} This is a failure case for \naq. While it correctly identifies a sink, this particular sink does not contain the bottle and the model fails to respond. 
    }
    \label{fig:qualitative}
    \vspace*{-0.1in}
\end{figure*}

\subsection{Performance analyses}
\label{sec:performance_analyses}
In the previous section, we verified the effectiveness of our approach through a careful comparison with recent state-of-the-art methods. We now ascertain the strengths and weaknesses of our approach through a series of quantitative studies and discuss qualitative results in~\cref{fig:qualitative}. For performing analysis-specific experiments, we adopt the EgoVLP + \naq method since it requires lower computational cost and time to train. \vspace{-0.05in} 

\begin{figure}[t]
    \centering
    \includegraphics[width=0.45\textwidth,clip,trim={0.0cm 9.5cm 13cm 0.0cm}]{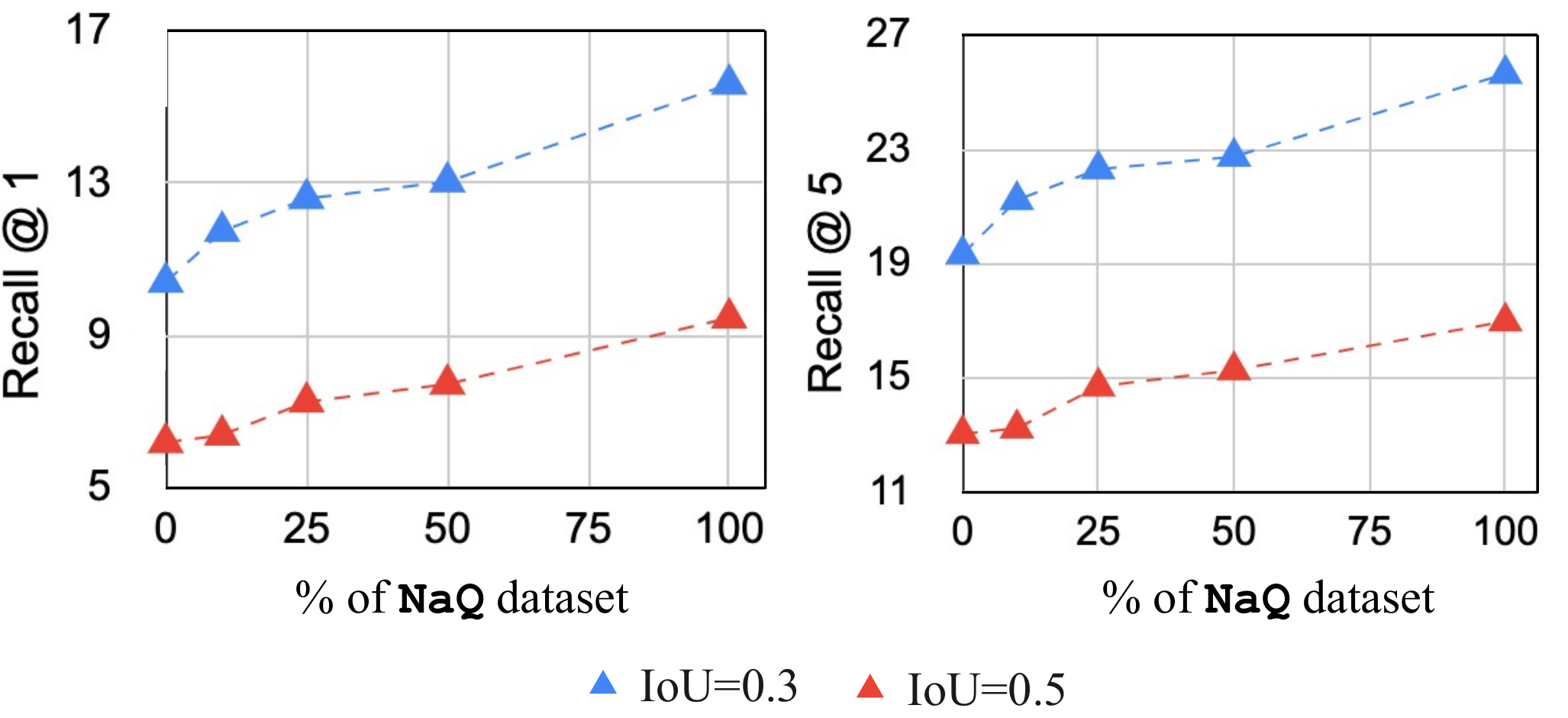}
    \vspace{-0.2cm}
    \caption{\textbf{Data scaling analysis.} We train EgoVLP + \naq using all NLQ and $k\%$ of \naq dataset ($k$ represented on the X-axis). NLQ performance scales linearly with the size of the \naq dataset.}
    \label{fig:scaling_chart}
    \vspace{-0.4cm}
\end{figure}

\quespara{(1) How does performance scale with narrations}{
One of the key benefits of using narrations for pretraining is that they are available on a large scale. We generated 850k narrations as queries for the NLQ task, which is two orders larger than the NLQ dataset containing 11.3k train queries. We now study performance scaling as a function of the amount of narrations used for training. For this, we additionally trained EgoVLP + \naq with $10\%$, $25\%$, $50\%$ of the narrations.~\cref{fig:scaling_chart} shows the results on NLQ (val). The $0\%$ performance represents EgoVLP and the $100\%$ performance represents the full EgoVLP + \naq reported in Tab.~\ref{tab:nlq-results}. When adding only $10\%$ of our \naq data, we already observe good improvements on all metrics. The performance continues to linearly scale as we add more narrations for \naq augmentation, confirming the utility of our paradigm. 
}\vspace{-0.05in}

\quespara{(2) What types of queries does \naq benefit}{
Next, we break down the NLQ performance across query types, i.e., the form of reasoning required by the query (e.g., \textit{where did I put object X? who did I talk to while doing activity Y?}). The NLQ dataset was created by providing an initial set of 13 query templates~\cite{grauman2022ego4d}. For reliable evaluation, we select 10 out of the 13 templates which contain 100 or more samples in the validation split, and report results in~\cref{tab:result_query_types} in supplementary. We observe that using \naq leads to significant improvements (marked in green) on 8/10 templates for at least 2/3 methods. However, it only has a limited impact for \textit{`Where is object X?'} and \textit{`In what location did I see X?'} queries. These queries may require explicit spatial understanding to achieve better performance.  Since all methods perform poorly on those queries and do not benefit from training on \naq, it hints at the need to incorporate better spatial understanding for video models. 
}\vspace{-0.05in}

\begin{table}[t]
\centering
\resizebox{0.47\textwidth}{!}{
\begin{tabular}{@{}lcccccc@{}}
\toprule
              &     \mctwo{High-shot}       &       \mctwo{Mid-shot}        &        \mctwo{Low-shot}        \\
Method        & \ftext{IoU=0.3} & \ftext{IoU=0.5}   &  \ftext{IoU=0.3}  &  \ftext{IoU=0.5}  & \ftext{IoU=0.3}&\ftext{IoU=0.5}\\\midrule
VSLNet        &      6.28   &      3.14     &      4.11     &      2.74 &      4.05      &     2.36      \\
\ rcol{+\naq}  & \tbg{9.72}  & \tbg{5.53}   & \tbg{11.42}   & \tbg{6.85}& \tbg{10.30}     &\tbg{5.57}     \\ \midrule
EgoVLP        &     13.15   &      7.17     &     10.20     &      5.63 &     10.64      &     5.91      \\
\rcol{+\naq}  &\tbg{16.59}  & \tbg{9.27}    &\tbg{16.13}    &\tbg{10.20}&\tbg{16.05}     &\tbg{10.30}    \\ \midrule
ReLER$^*$     &     17.07   &       9.95    &     17.88     &     10.73 &     13.36      &     8.29      \\
\rcol{+\naq}  &\tbg{21.24}  & \tbg{12.37}   &\tbg{21.60}    &\tbg{12.24}&\tbg{17.36}     &\tbg{10.29}    \\
\bottomrule
\end{tabular}
}
\vspace{-0.2cm}
\caption{\textbf{Performance breakdown across object types.} For object type queries, we categorize objects into low-shot, mid-shot, and high-shot objects based on their frequency of occurrence. We report the recall@1 metric at IoU=0.3 and IoU=0.5. We highlight cases where \naq \tbg{improves} recall over the baseline. 
}
\label{tab:result_object_types}
\vspace{-0.2cm}
\end{table}

\quespara{(3) Does \naq help respond about long-tail objects} 
The NLQ dataset has a long-tail of objects that are the subject of queries due to the sparse nature of NLQ annotations (1 query per 1.4 minutes of videos on average). However, since narrations are more densely annotated throughout the video (20+ narrations per minute), they contain rich information about objects that are rarely queried about. We therefore study if pretraining NLQ localization models with narrations can help respond to queries about long-tail objects. We divide objects from the NLQ train annotations into 3 types (as shown in~\cref{fig:long_tail}): \textbf{1. high-shot objects} which are queried more than 50 times (65 in total), \textbf{2. mid-shot objects} which are queried about 10 to 50 times (147 in total), and \textbf{3. low-shot objects} which are queried about between 2 to 10 times (967 in total). The results are in~\cref{tab:result_object_types}. Overall, we observe that \naq improves performance by a large margin in most cases, and has the biggest gains on mid-shot and low-shot objects. This indicates that using narrations as queries helps mitigate some of the biases in the NLQ data, and improves responses to queries about less-frequently occurring objects. \vspace{-0.05in}

\begin{figure}[t]
    \centering
    \includegraphics[width=0.47\textwidth,clip,trim={0 8.5cm 10.5cm 0}]{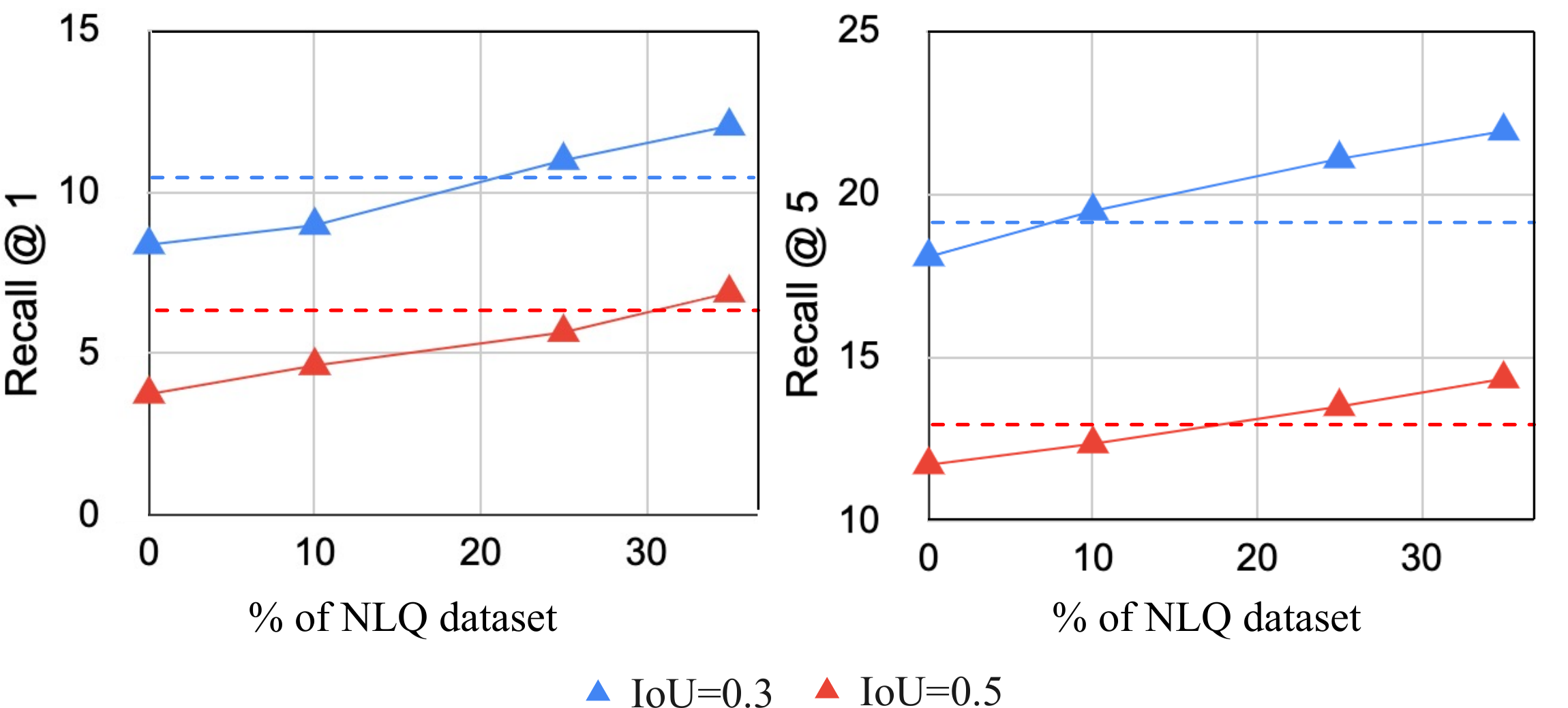}
    \vspace{-0.2cm}
    \caption{\textbf{Zero-shot and few-shot learning for NLQ}. We train EgoVLP + \naq using all \naq and $k\%$ of the NLQ train data ($k$ on the X-axis). The dotted horizontal lines represent the EgoVLP performance with $100\%$ NLQ and no \naq augmentation.
    }
    \label{fig:result_fewshot}
    \vspace{-0.5cm}
\end{figure}

\quespara{(4) Does \naq facilitate zero-shot / few-shot NLQ}{Considering that \naq enables better performance on long-tail objects, we next study whether it can facilitate zero-shot or few-shot learning for NLQ, i.e., given our large-scale \naq data and little to no NLQ task annotations, can we learn good NLQ models? We are first to study this to the best of our knowledge. We train EgoVLP + \naq method with all of \naq and $k\%$ of NLQ train data, where $k$ = $\{0, 10, 25, 35\}$.  $k=0$ represents the zero-shot case, and the rest represent few-shot learning. The results are in~\cref{fig:result_fewshot}. The triangles represent EgoVLP + \naq with $k\%$ NLQ data, and the horizontal line represents the EgoVLP baseline with no \naq data. It is interesting to observe that even with no NLQ data, the model performs well using \naq and competes closely with EgoVLP on the R@5 metrics. This generalization is facilitated by the use of BERT query-encoders that are pre-trained on large-scale text corpora. When we inject $10\%$ of the NLQ dataset, we get comparable or better performance on 2/4 metrics. At $25\%$ of NLQ data, it matches or outperforms EgoVLP on all metrics. Finally, at $35\%$, we outperform EgoVLP by a large margin. This study suggests that we can leverage large-scale free-form narrations using \naq to compensate for the lack of NLQ annotations. While these are not free to obtain, they are easier to annotate than NLQ and can also be used for various purposes other than the NLQ task itself~\cite{grauman2022ego4d}, meaning that many research directions are likely to continue investing in them.}

\section{Conclusions}

We propose Narrations-as-Queries, a simple data augmentation technique that dramatically improves state-of-the-art results on the Natural Language Queries task in the Ego4D Episodic Memory benchmark. Our key insight is to convert timestamped narrations in egocentric videos into natural language query annotations and use them as additional data for training NLQ localization models. To convert timestamped narrations into a form compatible with NLQ, we propose a temporal response jittering technique to convert a single timestamp into temporal windows. We perform experiments to demonstrate that our approach can be used as a simple plug-in to existing methods, massively improves multiple top methods for this task, and yields the very best performance to-date on the Ego4D NLQ benchmark. We hope that our approach serves as a useful tool for future research on this problem. Code, data, and models are available.

\section{Acknowledgements}
UT Austin is supported in part by the IFML NSF AI Institute and NSF CCRI.  KG is a paid as a researcher at Meta.

{\small
\bibliographystyle{ieee_fullname}
\bibliography{egbib}

\begin{thebibliography}{10}\itemsep=-1pt

\bibitem{abu2018will}
Yazan Abu~Farha, Alexander Richard, and Juergen Gall.
\newblock When will you do what?-anticipating temporal occurrences of
  activities.
\newblock In {\em Proceedings of the IEEE conference on computer vision and
  pattern recognition}, pages 5343--5352, 2018.

\bibitem{ashutosh2023hiervl}
Kumar Ashutosh, Rohit Girdhar, Lorenzo Torresani, and Kristen Grauman.
\newblock Hiervl: Learning hierarchical video-language embeddings.
\newblock In {\em Proceedings of the IEEE/CVF Conference on Computer Vision and
  Pattern Recognition}, 2023.

\bibitem{bain2021frozen}
Max Bain, Arsha Nagrani, G{\"u}l Varol, and Andrew Zisserman.
\newblock Frozen in time: A joint video and image encoder for end-to-end
  retrieval.
\newblock In {\em Proceedings of the IEEE/CVF International Conference on
  Computer Vision}, pages 1728--1738, 2021.

\bibitem{cai2018understanding}
Minjie Cai, Kris Kitani, and Yoichi Sato.
\newblock Understanding hand-object manipulation by modeling the contextual
  relationship between actions, grasp types and object attributes.
\newblock {\em arXiv preprint arXiv:1807.08254}, 2018.

\bibitem{chen2022internvideo}
Guo Chen, Sen Xing, Zhe Chen, Yi Wang, Kunchang Li, Yizhuo Li, Yi Liu, Jiahao
  Wang, Yin-Dong Zheng, Bingkun Huang, et~al.
\newblock Internvideo-ego4d: A pack of champion solutions to ego4d challenges.
\newblock {\em arXiv preprint arXiv:2211.09529}, 2022.

\bibitem{cui2022video}
Ran Cui, Tianwen Qian, Pai Peng, Elena Daskalaki, Jingjing Chen, Xiaowei Guo,
  Huyang Sun, and Yu-Gang Jiang.
\newblock Video moment retrieval from text queries via single frame annotation.
\newblock In {\em Proceedings of the 45th International ACM SIGIR Conference on
  Research and Development in Information Retrieval}, pages 1033--1043, 2022.

\bibitem{Damen2022RESCALING}
Dima Damen, Hazel Doughty, Giovanni~Maria Farinella, , Antonino Furnari, Jian
  Ma, Evangelos Kazakos, Davide Moltisanti, Jonathan Munro, Toby Perrett, Will
  Price, and Michael Wray.
\newblock Rescaling egocentric vision: Collection, pipeline and challenges for
  epic-kitchens-100.
\newblock {\em International Journal of Computer Vision (IJCV)}, 130:33–55,
  2022.

\bibitem{damen2014you}
Dima Damen, Teesid Leelasawassuk, Osian Haines, Andrew Calway, and Walterio~W
  Mayol-Cuevas.
\newblock You-do, i-learn: Discovering task relevant objects and their modes of
  interaction from multi-user egocentric video.
\newblock In {\em BMVC}, volume~2, page~3, 2014.

\bibitem{del2016summarization}
Ana~Garcia Del~Molino, Cheston Tan, Joo-Hwee Lim, and Ah-Hwee Tan.
\newblock Summarization of egocentric videos: A comprehensive survey.
\newblock {\em IEEE Transactions on Human-Machine Systems}, 47(1):65--76, 2016.

\bibitem{escorcia2019temporal}
Victor Escorcia, Mattia Soldan, Josef Sivic, Bernard Ghanem, and Bryan Russell.
\newblock Temporal localization of moments in video collections with natural
  language.
\newblock {\em arXiv preprint arXiv:1907.12763}, 2019.

\bibitem{fathi2011gtea}
Alireza Fathi, Xiaofeng Ren, and James~M. Rehg.
\newblock Learning to recognize objects in egocentric activities.
\newblock In {\em CVPR 2011}, pages 3281--3288, 2011.

\bibitem{feichtenhofer2019slowfast}
Christoph Feichtenhofer, Haoqi Fan, Jitendra Malik, and Kaiming He.
\newblock Slowfast networks for video recognition.
\newblock In {\em Proceedings of the IEEE/CVF international conference on
  computer vision}, pages 6202--6211, 2019.

\bibitem{furnari2020rolling}
Antonino Furnari and Giovanni~Maria Farinella.
\newblock Rolling-unrolling lstms for action anticipation from first-person
  video.
\newblock {\em IEEE transactions on pattern analysis and machine intelligence},
  43(11):4021--4036, 2020.

\bibitem{gao2017tall}
Jiyang Gao, Chen Sun, Zhenheng Yang, and Ram Nevatia.
\newblock Tall: Temporal activity localization via language query.
\newblock In {\em Proceedings of the IEEE international conference on computer
  vision}, pages 5267--5275, 2017.

\bibitem{girdhar2021anticipative}
Rohit Girdhar and Kristen Grauman.
\newblock Anticipative video transformer.
\newblock In {\em Proceedings of the IEEE/CVF International Conference on
  Computer Vision}, pages 13505--13515, 2021.

\bibitem{grauman2022ego4d}
Kristen Grauman, Andrew Westbury, Eugene Byrne, Zachary Chavis, Antonino
  Furnari, Rohit Girdhar, Jackson Hamburger, Hao Jiang, Miao Liu, Xingyu Liu,
  et~al.
\newblock Ego4d: Around the world in 3,000 hours of egocentric video.
\newblock In {\em Proceedings of the IEEE/CVF Conference on Computer Vision and
  Pattern Recognition}, pages 18995--19012, 2022.

\bibitem{hendricks2018localizing}
Lisa~Anne Hendricks, Oliver Wang, Eli Shechtman, Josef Sivic, Trevor Darrell,
  and Bryan Russell.
\newblock Localizing moments in video with temporal language.
\newblock In {\em EMNLP}, 2018.

\bibitem{hou2022efficient}
Zhijian Hou, Wanjun Zhong, Lei Ji, Difei Gao, Kun Yan, Wing-Kwong Chan,
  Chong-Wah Ngo, Zheng Shou, and Nan Duan.
\newblock An efficient coarse-to-fine alignment framework@ ego4d natural
  language queries challenge 2022.
\newblock {\em arXiv preprint arXiv:2211.08776}, 2022.

\bibitem{kazakos2019epic}
Evangelos Kazakos, Arsha Nagrani, Andrew Zisserman, and Dima Damen.
\newblock Epic-fusion: Audio-visual temporal binding for egocentric action
  recognition.
\newblock In {\em Proceedings of the IEEE/CVF International Conference on
  Computer Vision}, pages 5492--5501, 2019.

\bibitem{krishna2017dense}
Ranjay Krishna, Kenji Hata, Frederic Ren, Li Fei-Fei, and Juan Carlos~Niebles.
\newblock Dense-captioning events in videos.
\newblock In {\em Proceedings of the IEEE international conference on computer
  vision}, pages 706--715, 2017.

\bibitem{yongjae-ijcv2015}
Y~J. Lee and K. Grauman.
\newblock Predicting important objects for egocentric video summarization.
\newblock {\em International Journal on Computer Vision}, 2015.

\bibitem{li2020hero}
Linjie Li, Yen-Chun Chen, Yu Cheng, Zhe Gan, Licheng Yu, and Jingjing Liu.
\newblock Hero: Hierarchical encoder for video+ language omni-representation
  pre-training.
\newblock In {\em Proceedings of the 2020 Conference on Empirical Methods in
  Natural Language Processing (EMNLP)}, pages 2046--2065, 2020.

\bibitem{lin2022egocentric}
Kevin~Qinghong Lin, Alex~Jinpeng Wang, Mattia Soldan, Michael Wray, Rui Yan,
  Eric~Zhongcong Xu, Difei Gao, Rongcheng Tu, Wenzhe Zhao, Weijie Kong, et~al.
\newblock Egocentric video-language pretraining.
\newblock {\em arXiv preprint arXiv:2206.01670}, 2022.

\bibitem{liu2022reler}
Naiyuan Liu, Xiaohan Wang, Xiaobo Li, Yi Yang, and Yueting Zhuang.
\newblock Reler@ zju-alibaba submission to the ego4d natural language queries
  challenge 2022.
\newblock {\em arXiv preprint arXiv:2207.00383}, 2022.

\bibitem{miech2020end}
Antoine Miech, Jean-Baptiste Alayrac, Lucas Smaira, Ivan Laptev, Josef Sivic,
  and Andrew Zisserman.
\newblock End-to-end learning of visual representations from uncurated
  instructional videos.
\newblock In {\em Proceedings of the IEEE/CVF Conference on Computer Vision and
  Pattern Recognition}, pages 9879--9889, 2020.

\bibitem{miech2019howto100m}
Antoine Miech, Dimitri Zhukov, Jean-Baptiste Alayrac, Makarand Tapaswi, Ivan
  Laptev, and Josef Sivic.
\newblock Howto100m: Learning a text-video embedding by watching hundred
  million narrated video clips.
\newblock In {\em Proceedings of the IEEE/CVF International Conference on
  Computer Vision}, pages 2630--2640, 2019.

\bibitem{mo2022simple}
Sicheng Mo, Fangzhou Mu, and Yin Li.
\newblock A simple transformer-based model for ego4d natural language queries
  challenge.
\newblock {\em arXiv preprint arXiv:2211.08704}, 2022.

\bibitem{nair2022r3m}
Suraj Nair, Aravind Rajeswaran, Vikash Kumar, Chelsea Finn, and Abhinav Gupta.
\newblock R3m: A universal visual representation for robot manipulation.
\newblock {\em arXiv preprint arXiv:2203.12601}, 2022.

\bibitem{radford2021learning}
Alec Radford, Jong~Wook Kim, Chris Hallacy, Aditya Ramesh, Gabriel Goh,
  Sandhini Agarwal, Girish Sastry, Amanda Askell, Pamela Mishkin, Jack Clark,
  et~al.
\newblock Learning transferable visual models from natural language
  supervision.
\newblock In {\em International Conference on Machine Learning}, pages
  8748--8763. PMLR, 2021.

\bibitem{rohrbach2014coherent}
Anna Rohrbach, Marcus Rohrbach, Wei Qiu, Annemarie Friedrich, Manfred Pinkal,
  and Bernt Schiele.
\newblock Coherent multi-sentence video description with variable level of
  detail.
\newblock In {\em Pattern Recognition: 36th German Conference, GCPR 2014,
  M{\"u}nster, Germany, September 2-5, 2014, Proceedings 36}, pages 184--195.
  Springer, 2014.

\bibitem{rohrbach2017movie}
Anna Rohrbach, Atousa Torabi, Marcus Rohrbach, Niket Tandon, Christopher Pal,
  Hugo Larochelle, Aaron Courville, and Bernt Schiele.
\newblock Movie description.
\newblock {\em International Journal of Computer Vision}, 123(1):94--120, 2017.

\bibitem{seo2016bidirectional}
Minjoon Seo, Aniruddha Kembhavi, Ali Farhadi, and Hannaneh Hajishirzi.
\newblock Bidirectional attention flow for machine comprehension.
\newblock {\em arXiv preprint arXiv:1611.01603}, 2016.

\bibitem{sun2019videobert}
Chen Sun, Austin Myers, Carl Vondrick, Kevin Murphy, and Cordelia Schmid.
\newblock Videobert: A joint model for video and language representation
  learning.
\newblock In {\em Proceedings of the IEEE/CVF International Conference on
  Computer Vision}, pages 7464--7473, 2019.

\bibitem{xu2017video}
Dejing Xu, Zhou Zhao, Jun Xiao, Fei Wu, Hanwang Zhang, Xiangnan He, and Yueting
  Zhuang.
\newblock Video question answering via gradually refined attention over
  appearance and motion.
\newblock In {\em Proceedings of the 25th ACM international conference on
  Multimedia}, pages 1645--1653, 2017.

\bibitem{xu2021vlm}
Hu Xu, Gargi Ghosh, Po-Yao Huang, Prahal Arora, Masoumeh Aminzadeh, Christoph
  Feichtenhofer, Florian Metze, and Luke Zettlemoyer.
\newblock Vlm: Task-agnostic video-language model pre-training for video
  understanding.
\newblock In {\em Findings of the Association for Computational Linguistics:
  ACL-IJCNLP 2021}, pages 4227--4239, 2021.

\bibitem{xu2022point}
Zhe Xu, Kun Wei, Xu Yang, and Cheng Deng.
\newblock Point-supervised video temporal grounding.
\newblock {\em IEEE Transactions on Multimedia}, pages 1--11, 2022.

\bibitem{yang2021just}
Antoine Yang, Antoine Miech, Josef Sivic, Ivan Laptev, and Cordelia Schmid.
\newblock Just ask: Learning to answer questions from millions of narrated
  videos.
\newblock In {\em Proceedings of the IEEE/CVF International Conference on
  Computer Vision}, pages 1686--1697, 2021.

\bibitem{zhang2020span}
Hao Zhang, Aixin Sun, Wei Jing, and Joey~Tianyi Zhou.
\newblock Span-based localizing network for natural language video
  localization.
\newblock In {\em Proceedings of the 58th Annual Meeting of the Association for
  Computational Linguistics}, pages 6543--6554, 2020.

\bibitem{zhang2020learning}
Songyang Zhang, Houwen Peng, Jianlong Fu, and Jiebo Luo.
\newblock Learning 2d temporal adjacent networks for moment localization with
  natural language.
\newblock In {\em Proceedings of the AAAI Conference on Artificial
  Intelligence}, volume~34, pages 12870--12877, 2020.

\bibitem{zheng2022exploring}
Sipeng Zheng, Qi Zhang, Bei Liu, Qin Jin, and Jianlong Fu.
\newblock Exploring anchor-based detection for ego4d natural language query.
\newblock {\em arXiv preprint arXiv:2208.05375}, 2022.

\bibitem{zhou2018end}
Luowei Zhou, Yingbo Zhou, Jason~J Corso, Richard Socher, and Caiming Xiong.
\newblock End-to-end dense video captioning with masked transformer.
\newblock In {\em Proceedings of the IEEE conference on computer vision and
  pattern recognition}, pages 8739--8748, 2018.

\bibitem{zhou2015temporal}
Yipin Zhou and Tamara~L Berg.
\newblock Temporal perception and prediction in ego-centric video.
\newblock In {\em Proceedings of the IEEE International Conference on Computer
  Vision}, pages 4498--4506, 2015.

\end{thebibliography}
}

\newpage
\clearpage

\setcounter{section}{0}
\setcounter{figure}{0}
\setcounter{table}{0}
\renewcommand{\thesection}{S\arabic{section}}
\renewcommand{\thetable}{S\arabic{table}}
\renewcommand{\thefigure}{S\arabic{figure}}

\section*{Supplementary Materials}
We now provide additional information about our experimental settings, and qualitative and quantitative analyses to support our experiments in the main paper.

\section{Implementation details}
\label{suppsec:implementation}

We perform joint \naq + NLQ training with a large batch sizes and high learning rates for accelerated convergence. For VSLNet and EgoVLP methods, we use a batch size of 2048 and initial learning rate of $0.001$ on 2 A40 GPUs with a memory size of 46GB per GPU. For ReLER$^*$, we use a batch size of 1536 and an initial learning rate of $0.001$ on 8 A40 GPUs since it has larger memory and compute requirements. We train each method for up to 200 epochs on \naq + NLQ training data, and then finetune them for up to 30 epochs on NLQ training data alone with a lower learning rate. We found finetuning to be unnecessary for VSLNet. For EgoVLP, we finetuned with the original hyperparameter settings from~\cite{lin2022egocentric} and a learning rate of $0.00001$. For ReLER$^*$, we finetuned with the original hyperparameter setting from~\cite{liu2022reler} and a learning rate of $0.0001$. We perform early stopping in each case using the performance on NLQ validation split. 

For temporal random jittering (TRJ), we performed a grid search with the expansion factor values $S$=$\{2.5, 5.0, 10.0, 20.0\}$. We found $S$=$5.0$ to work best for EgoVLP based on the NLQ validation performance (see~\cref{supptab:trj_hyperparam}). Similarly, we found $S$=$2.5$ to work best for ReLER* and VSLNet.

\begin{table}[t]
\centering
\resizebox{0.45\textwidth}{!}{%
\begin{tabular}{@{}ccccc@{}}
\toprule
               &   $S=2.5$   &    $S=5.0$    & $S=10.0$     & $S=20.0$     \\ 
R@1, IoU=0.5   &    8.64     &  \tb{9.46}    &     8.08     &     8.47     \\
\bottomrule
\end{tabular}
}
\caption{Varying TRJ scale $S$ with EgoVLP (and \naq stage 1).}
\label{supptab:trj_hyperparam}
\end{table}

\section{Long-tail of objects in NLQ}
\label{suppsec:longtail}

\begin{figure}[t]
    \centering
    \includegraphics[width=0.48\textwidth,clip,trim={0 8.0cm 0 0}]{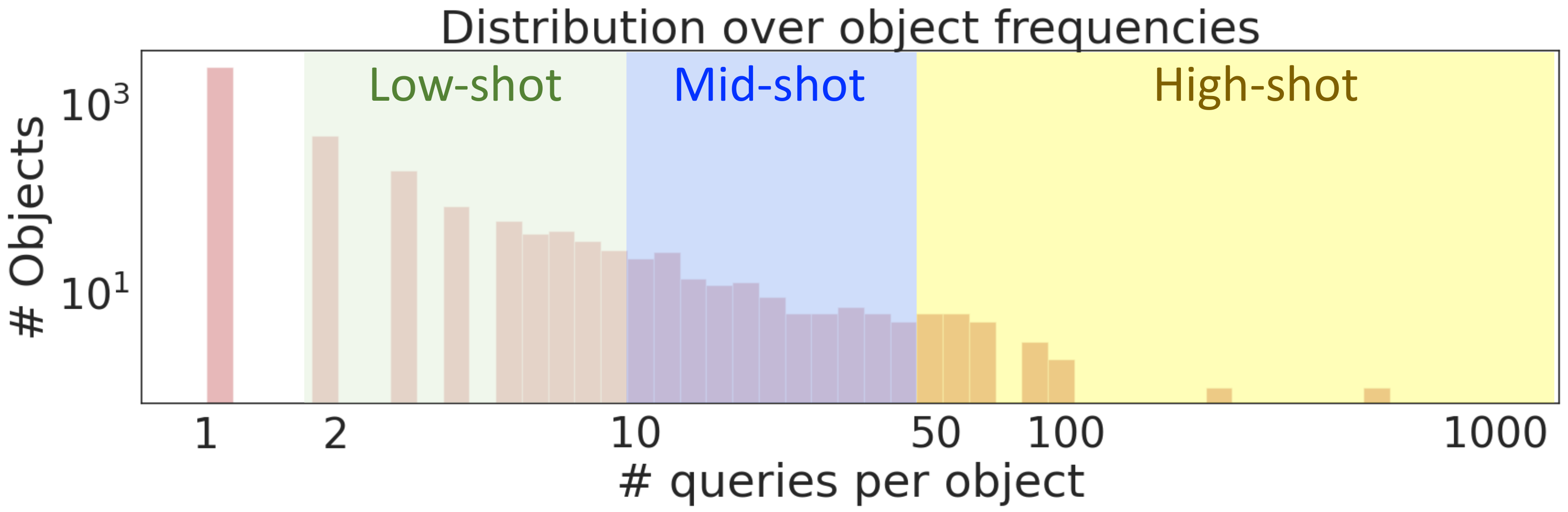}
    \caption{Long-tail of objects in NLQ.}
    \label{fig:long_tail}
\end{figure}

\cref{fig:long_tail} shows the long-tail of objects queried about in NLQ, and the split of low-shot, mid-shot, and high-shot objects used in~\cref{sec:performance_analyses}. Note that for a given point $x$ on X-axis, the Y-axis shows the number of objects that have $x$ queries in the NLQ train dataset. For example, there are more than 1000 objects with only 1 training sample.

\section{Ablation study for Temporal Response Jittering}
\label{suppsec:ablations}

We study the impact of using temporal response jittering (TRJ) described in~\cref{eqn:trj}. 
In~\cref{tab:ablation-nlq-results}, we measure the performance of using \naq with and without TRJ, where not using TRJ implies that the seed temporal window from~\cref{eqn:seed}
is used.
Overall, we observe a consistent improvement of up to $0.80$ in R@1 metrics and $1.62$ in R@5 metrics. This indicates that TRJ is able to address the limitations of the seed temporal window.

\begin{table}[t]
\centering
\resizebox{0.48\textwidth}{!}{%
\begin{tabular}{@{}lccccc@{}}
\toprule
                                          &                     & \multicolumn{2}{c}{IoU=0.3}     & \multicolumn{2}{c}{IoU=0.5} \\ 
\multicolumn{1}{c}{Method}                & {\small TRJ} &       R@1      &     R@5        &     R@1      &    R@5       \\ \midrule
VSLNet + \naq                             &  \xmark      &       9.80     &     18.05      &    5.27      &    11.05     \\
VSLNet + \naq                             &  \cmark      &   \tb{10.26}   & \tb{19.01}     &\tb{5.81}     &\tb{12.67}    \\
\multicolumn{1}{r}{\small absolute gain}  &              & \tbg{+0.46}    & \tbg{+0.96}    &\tbg{+0.54}   &\tbg{+1.62}    \\ \midrule
EgoVLP + \naq                             &  \xmark      &       15.25    &     26.15      &    9.12      &    17.63     \\
EgoVLP + \naq                             &  \cmark      &   \tb{15.90}   & \tb{26.38}     &\tb{9.46}     &\tb{17.80}    \\
\multicolumn{1}{r}{\small absolute gain}  &              & \tbg{+0.65}    &\tbg{+0.23}     &\tbg{+0.34}   &\tbg{+0.17}    \\ \midrule
ReLER$^*$ + \naq                          &  \xmark      &       18.51    &     23.23      &    11.36     &    15.44     \\
ReLER$^*$ + \naq                          &  \cmark      &   \tb{19.31}   & \tb{23.62}     &\tb{11.59}    &\tb{15.75}    \\
\multicolumn{1}{r}{\small absolute gain}  &              & \tbg{+0.80}    &\tbg{+0.39}     &\tbg{+0.23}   &\tbg{+0.31}    \\
\bottomrule
\end{tabular}%
}
\vspace{-0.25cm}
\caption{Ablation study of temporal random jittering (TRJ).}
\label{tab:ablation-nlq-results}
\vspace{-0.25cm}
\end{table}

\begin{table*}[t]
\centering
\resizebox{\textwidth}{!}{
\begin{tabular}{@{}l|P{1.5cm}P{1.5cm}P{0.75cm}P{1.5cm}P{1.4cm}P{1.5cm}P{1.1cm}P{1.0cm}P{0.80cm}|P{1.8cm}@{}} & \multicolumn{9}{c|}{Object / place queries} & \multicolumn{1}{c}{People queries} \\ \cmidrule{2-11}
\multicolumn{1}{c|}{Method} & {\small Where is X before/after Y?} & {\small Where did I put X?} & {\small Where is X?} & {\small What did I put in X?} & {\small How many X's?} & {\small In what location did I see X?} & {\small What X did I Y?} & {\small What X is Y?} & {\small State?} & {\small Who did I interact with during Y?} \\ \midrule
VSLNet                      &   2.04     &    1.07     &    3.38     &    3.23       &     4.91      &     2.60   &     3.64     &     2.34   &     3.07    &     3.26     \\
\multicolumn{1}{r|}{+\naq}  & \tbg{6.62} &\tbg{3.58}   &    3.14     &\tbg{5.76}     & \tbg{10.18}   &     2.60   &\tbg{8.61}    &\tbg{5.86}  &\tbg{8.59}   &\tbg{6.52}    \\ \midrule
EgoVLP                      &   5.77     &    3.58     &    4.11     &     9.45      &     9.82      &     2.97   &      7.62    &     5.08   &      7.98   &     8.70     \\
\multicolumn{1}{r|}{+\naq}  &\tbg{10.70} &\tbg{6.44}   &\tbg{4.83}   &\tbg{13.13}    &\tbg{15.79}    &     2.60   &\tbg{11.59}   &\tbg{7.03}  &\tbg{12.88}  &\tbg{13.04}   \\\midrule
ReLER*                      &    9.63    &    6.87     &    5.82     &    10.71      &    14.33      &     5.46   &     11.54    &     6.54   &     10.12   &     4.90     \\
\multicolumn{1}{r|}{+\naq}  &\tbg{13.98} &\tbg{11.34}  &    6.04     &\tbg{12.39}    &\tbg{21.00}    & \tbr{4.78} &\tbg{15.38}   &     6.54   &\tbg{14.29}  &\tbg{7.84}    \\ \bottomrule
\end{tabular}
}
\caption{\textbf{Performance over NLQ query types.} We report recall@1 at IoU=0.5. We include query types with $\ge100$ val samples. We highlight cases where \naq \tbg{improves} recall by more than $0.5$ points. 
}
\label{tab:result_query_types}
\vspace*{-0.5cm}
\end{table*}

\section{Impact of two-stage \naq training}

As we stated in~\cref{sec:expt_setup}, we train models using \naq augmentation in two stages. In the first stage, we jointly train models on the combined \naq and NLQ dataset with large batch training. In the second stage, we finetune the models on only the NLQ dataset with standard training. In~\cref{supptab:naq_stages}, we study the impact of each stage of training. The first stage helps the most. Stage 2 is not critical, but useful nonetheless (except for VSLNet).\\

\begin{table}[t]
\centering
\resizebox{0.46\textwidth}{!}{%
\begin{tabular}{@{}ccccc@{}}
\toprule
                         & \multicolumn{2}{c}{IoU = 0.3} & \multicolumn{2}{c}{IoU=0.5} \\
Method                   & R@1           & R@5           & R@1          & R@5          \\ \midrule
VSLNet                   &      5.21     &     11.19     &     2.78     &      6.72    \\
VSLNet + \naq stage 1    & \tb{10.26}    & \tb{19.01}    & \tb{5.81}    & \tb{12.67}    \\
VSLNet + \naq stage 1,2  &      9.34     &     17.66     &     5.29     &     11.85     \\ \midrule
EgoVLP                   &     10.40     &     19.33     &     6.18     &     13.03     \\
EgoVLP + \naq stage 1    &     15.61     &     25.64     & \tb{9.46}    &     16.97     \\
EgoVLP + \naq stage 1,2  & \tb{15.90}    & \tb{26.38}    & \tb{9.46}    & \tb{17.80}    \\ \midrule
ReLER*                   &     14.66     &     17.84     &     8.67     &     11.54     \\
ReLER* + \naq stage 1    &     18.28     &     22.95     &     10.38    &     14.82     \\
ReLER* + \naq stages 1,2 & \tb{19.31}    & \tb{23.62}    & \tb{11.59}   & \tb{15.75}    \\ \bottomrule
\end{tabular}
}
\caption{Impact of two-stage \naq training.}
\label{supptab:naq_stages}
\end{table}

\begin{table*}[t]   
\centering
\resizebox{\textwidth}{!}{%
\begin{tabular}{@{}cc|P{1.5cm}P{1.5cm}P{0.75cm}P{1.5cm}P{1.4cm}P{1.5cm}P{1.1cm}P{1.0cm}P{0.80cm}|P{1.8cm}@{}} & & \multicolumn{9}{c|}{Object / place queries} & \multicolumn{1}{c}{People queries} \\ \cmidrule{3-12}
\% NLQ & \% \naqb & {\small Where is X before/after Y?} & {\small Where did I put X?} & {\small Where is X?} & {\small What did I put in X?} & {\small How many X's?} & {\small In what location did I see X?} & {\small What X did I Y?} & {\small What X is Y?} & {\small State?} & {\small Who did I interact with during Y?} \\ \midrule
 100   &    0    &   5.77     &    3.58     &    4.11     &     9.45      &     9.82      &     2.97   &      7.62    &     5.08   &      7.98   &     8.70      \\\midrule
   0   &  100    &   3.88     &    3.83     &    2.68     &     2.31      &     5.00      &     1.37   &      4.17    &     5.23   &      7.14   &     2.94       \\
  10   &  100    &   7.45     &    5.11     &    2.46     &     4.62      &     6.00      &     1.02   &      3.53    &     4.90   &      5.95   &     3.92       \\
  25   &  100    &   9.63     &    4.63     &    3.13     &     5.46      &     7.67      &     1.37   &      4.81    &     5.23   &      5.95   &     3.92       \\
  35   &  100    &   8.54     &    4.95     &    3.58     &     7.14      &    13.33      &     4.10   &      7.37    &     6.54   &      7.74   &     4.90       \\
 100   &  100    &  10.70     &    6.44     &    4.83     &     13.13     &    15.79      &     2.60   &     11.59    &     7.03   &     12.88   &    13.04       \\\bottomrule
\end{tabular}%
}   
\caption{\textbf{Few-shot analysis.} We split the few-shot results from Fig.~6 in the main paper across the various query templates. We report recall@1 at IoU=0.5.  The first two columns show the percentage of the NLQ and \naq data used for training. For example, the first row with 100\% NLQ and 0\% \naq is the baseline, the second row with 0\% NLQ and 100\% \naq is our zero-shot setting, and so on. 
}
\label{tab:zero_shot_analysis}
\vspace*{-0.5cm}
\end{table*}

\section{Few-shot analysis}
\label{suppsec:fewshot}

We perform a more detailed analysis of the few-shot performance discussed in~\cref{sec:performance_analyses} and~\cref{fig:result_fewshot}.
Specifically, we analyze the zero-/few-shot performance across the various query templates in~\cref{tab:zero_shot_analysis}. When tested zero-shot, \naq already competes with or outperforms the baseline on object/place templates such as `where did I put X?', `what X is Y?', and `object state'. As we inject NLQ data into \naq training, the performance improves steadily on the remaining templates, and outperforms the baseline on 9/10 templates. 

\section{Qualitative examples}
\label{suppsec:qualitative}

In \texttt{\small supplementary.html} shared \href{https://utexas.box.com/s/6b17fmxt89zqbnpvhoytdm3pyg5m1xey}{here}, we link to qualitative videos for the following:
\begin{itemize}
    \item Comparing annotations for NLQ vs. Narrations
    \item NaQ benefits performance on most query templates
    \item NaQ benefits performance on queries about long-tail objects
    \item NaQ facilitates zero-shot NLQ
\end{itemize}

\end{document}